\documentclass{article}

\usepackage[preprint]{neurips_2021}

\usepackage[utf8]{inputenc} 
\usepackage[T1]{fontenc}    
\usepackage{hyperref}       
\usepackage{url}            
\usepackage{booktabs}       
\usepackage{amsfonts}       
\usepackage{nicefrac}       
\usepackage{microtype}      
\usepackage{xcolor}         

\usepackage{amsmath,amsfonts,amssymb}
\usepackage[ruled]{algorithm2e}
\usepackage{graphicx}
\usepackage{caption}
\usepackage{subcaption}

\title{Learning Agent State Online \\ with Recurrent Generate-and-Test}

\author{%
  Amir Samani \\
  Department of Computing Science\\
  University of Alberta\\
  \texttt{samani@ualberta.ca} \\
  \And
  Richard S. Sutton\\
  Department of Computing Science\\
  University of Alberta\\
  \texttt{rsutton@ualberta.ca} \\
}

\begin{document}

\maketitle

\begin{abstract}
Learning continually and online from a continuous stream of data is challenging, especially for a reinforcement learning agent with sequential data. When the environment only provides \textit{observations} giving partial information about the state of the environment, the agent must learn the \textit{agent state} based on the data stream of experience. We refer to the state learned directly from the data stream of experience as the agent state. Recurrent neural networks can learn the agent state, but the training methods are computationally expensive and sensitive to the hyper-parameters, making them unideal for online learning. This work introduces methods based on the \textit{generate-and-test} approach to learn the agent state. A generate-and-test algorithm searches for state features by generating features and testing their usefulness. In this process, features useful for the agent's performance on the task are preserved, and the least useful features get replaced with newly generated features. We study the effectiveness of our methods on two online multi-step prediction problems. The first problem, \textit{trace conditioning}, focuses on the agent's ability to remember a cue for a prediction multiple steps into the future. In the second problem, \textit{trace patterning}, the agent needs to learn patterns in the observation signals and remember them for future predictions. We show that our proposed methods can effectively learn the agent state online and produce accurate predictions.
\end{abstract}
\section{Introduction}
\label{sec:intro}
Online continual learning refers to learning from a never-ending data stream without reusing past data points. A reinforcement learning agent interacts with its environment by performing actions and receiving observations. This interaction results in the agent's data stream of experience. In many cases of interest, the agent does not have access to the underlying state of the environment and, when interacting with the environment, only receives observations that provide partial information. One of the challenges that the agent has to overcome is representing its state based on the data stream of experience, which we call the agent state. Similarly, natural intelligent agents only receive limited information about the state of the environment; for instance, objects are not visible in the dark or when distant. Nevertheless, studies in classical conditioning show that animals can make accurate multi-step predictions, suggesting that animals make representations that summarize their interaction with the environment. 

In reinforcement learning, learning the state is essential to the agent as the state is used in the agent's policy, value functions, and the environmental model. Historically, domain experts designed the state based on their knowledge of the environment. Relying on human input to learn the state is at odds with a significant strength of reinforcement learning which is the ability to learn directly from the data stream of experience. Therefore, we would like online algorithms to learn the agent state using the data stream of experience.

Modern deep learning algorithms based on gradient descent, such as real-time recurrent learning (RTRL) and backpropagation through time (BPTT), learn the agent state based on the data stream of experience. However, these algorithms are expensive in memory and computation and sensitive to the selection of their hyper-parameters. We want learning algorithms that are inexpensive in terms of computation and memory, which naturally fit with reinforcement learning algorithms for prediction and control. 

A simple approach for learning the agent state is the generate-and-test. \citet{Mahmood13GenTest} propose the generate-and-test approach to search for features that improve the performance. This search method for finding features generates candidate features and tests them for their utility on a given task. A \textit{generator} would generate features, and a \textit{tester} preserves the more useful features and deletes the least useful ones. \citet{Mahmood13GenTest} apply the generate-and-test to a synthetic supervised learning problem with a feed-forward network, and we would like to extend generate-and-test to reinforcement learning and sequential data and thus with a recurrent network to learn the agent state for a reinforcement learning agent.

Classical conditioning experiments on animals such as dogs and rabbits have shown that an unconditioned stimulus (US) such as food can get associated with a conditioned stimulus (CS) such as a bell after several pairings \citep{pavlov1927conditioned}. For instance, after multiple pairings, a dog would start salivating after hearing the bell in anticipation of the food. These experiments suggest that animals can make accurate multi-step predictions. Inspired by these experiments on animals, \cite{Rafiee21Testbed} introduce benchmarks for partially observable multi-step online prediction problems. In this paper, we focus on the trace conditioning problem and trace patterning problem. In the trace conditioning problem, the agent must predict the US multiple time steps in the future using a CS, similar to predicting the arrival of food using the sound of the bell. In trace patterning, the agent still needs to predict the US multiple time step in the future. However, there are several CSs, and only a specific configuration of the CSs results in the arrival of the US. For instance, the dog would receive the food if only the bell sound and the light were present.

\begin{figure}[ht]
\begin{center}
\includegraphics[width=14cm]{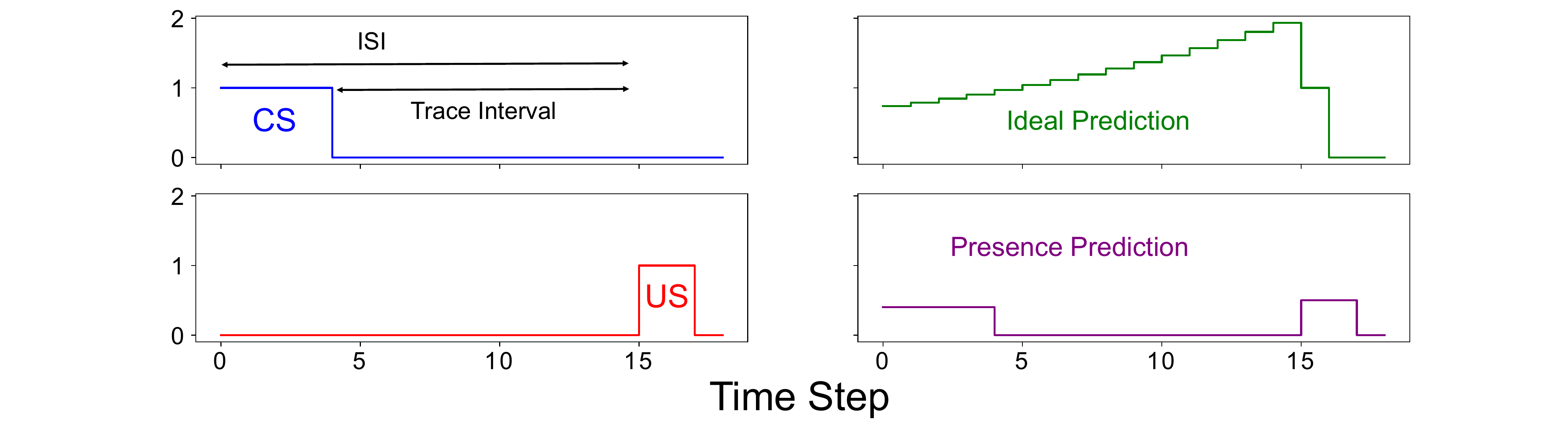}
\caption{An example of the trace conditioning problem in which the CS is active for 4 time steps (top right) the US arrives 15 time steps after the onset of the CS (bottom right). The time from the onset of the CS to the onset of the US is called the inter-stimulus interval (ISI). The time from the offset of the CS to the onset of the US is called the trace Interval. During the trace interval, there are no relevant observation signals available to the agent. The agent needs to learn the agent state to predict the arrival of the US accurately. Studies have shown that the predictions made by the animals are similar to the discounted return (top right). A simple form of representing the state, the presence representation, uses the observation as the agent state thus can not make accurate predictions, especially when the trace interval gap is long (bottom right). }
\label{fig:cc-example}
\end{center}
\end{figure}
This paper proposes two online generators and a simple tester for learning the agent state. First, we discuss the architecture for representing the agent state. Second, we propose the deep trace generator that allows the agent to fill the trace interval gap by making features that hold fading memory of features and observation signals. We study the effectiveness of the deep trace generator on the trace conditioning problem. We show that the agent can fill the trace interval gap using the feature generated by the deep trace generator. Finally, we enable the agent to make features representing a non-linear configuration of the stimuli using our second generator, the imprinting generator. The imprinting generator makes features that respond to a particular configuration in the observation signals. We show that combining the deep trace generator and the imprinting generator allows the agent to make accurate predictions in the trace patterning problem.

\section{Agent State Architecture}
\label{sec:agent-state}
The data stream of experience for a reinforcement learning agent contains the history of all the interactions between the agent and the environment---actions performed by the agent and the observations received from the environment. We denote the action at time step $t$ by $\mathbf{a}_t  \in \mathbb{R}^d$ and the observation at time $t$ by $\mathbf{o}_t  \in \mathbb{R}^m$. The data stream of experience is the sequence
\[\mathbf{a}_0, \mathbf{o}_1, \mathbf{a}_1, \mathbf{o}_2, \mathbf{a}_2, \mathbf{o}_3, ...,\mathbf{a}_{t-1}, \mathbf{o}_{t}, ...\]
going forever for the life the reinforcement learning agent. Storing the whole sequence and using it as the agent state is computationally impractical since the length of the sequence grows with time. Let us denote the agent state at time step $t$ as $\mathbf{s}_t  \in \mathbb{R}^n$. We prefer the agent state to be updated incrementally based on the previous agent state $\mathbf{s}_{t-1}$ and the most recent observation $\mathbf{o}_t$ and action $\mathbf{a}_{t-1}$. We call this update function \textit{state-update} function and denote it by $u$:
\[\mathbf{s}_t = u(\mathbf{s}_{t-1}, \mathbf{o}_t, \mathbf{a}_{t-1}).\]

We want to learn the agent state by learning the state-update function $u$. We consider the case in which the agent state consists of features. Since the features are learned, we may talk about the usefulness of each feature. We can compare features based on their usefulness for predicting or controlling the data stream of experience. Our method to learning the state is based on the generate-and-test approach. We search for more useful features by generating features and putting them through testing. As the agent performs on the original task, the generator makes new features. The agent may use these features for prediction or controlling its data stream of experience. The tester preserves features that the agent found to be useful and eliminates the least useful features. With the deletion of the least useful feature, the generator can then make more features. Although similar ideas have been studied before, our approach extends representation search to partially observable sequential multi-step prediction setting for learning the agent state.

 Each feature $s_t^i \in \mathbb{R}$ may be connected to $x_{t}^{j}$ with the weight of $V_{t}^{i,j}$ where $\mathbf{x}_{t}=[\mathbf{s}_{t-1},\mathbf{o}_t,\mathbf{a}_{t-1}] \in \mathbb{R}^{m+n+d}$ is the concatenation of previous time step state $\mathbf{s}_{t-1}$ and the most recent observation $\mathbf{o}_{t}$ and action $\mathbf{a}_{t-1}$. The feature $s_t^i$ is then computed as follows:
\[s_t^i = \sum_{j=1}^{m+n+d} V_{t}^{i,j}x_{t}^{j}.\]

In the case of a single scalar prediction, the prediction $y_t \in \mathbb{R}$ at time step $t$ is connected to $f_t^k$ with the weight of $w_t^k$ where $\mathbf{f}_t=[s_{t},o_t,a_{t-1}] \in \mathbb{R}^{m+n+d}$ is the concatenation current time step state $\mathbf{s}_{t}$ and the most recent observation $\mathbf{o}_{t}$ and action $\mathbf{a}_{t-1}$. The final prediction $y_t$ is compute as follows: 
\[y_t = \sum_{k=1}^{m+n+d} w_{t}^k f_t^k = {\mathbf f}_{t}^\intercal{\mathbf w}_t .\]

\begin{figure}[ht]
\begin{center}
\includegraphics[width=10cm]{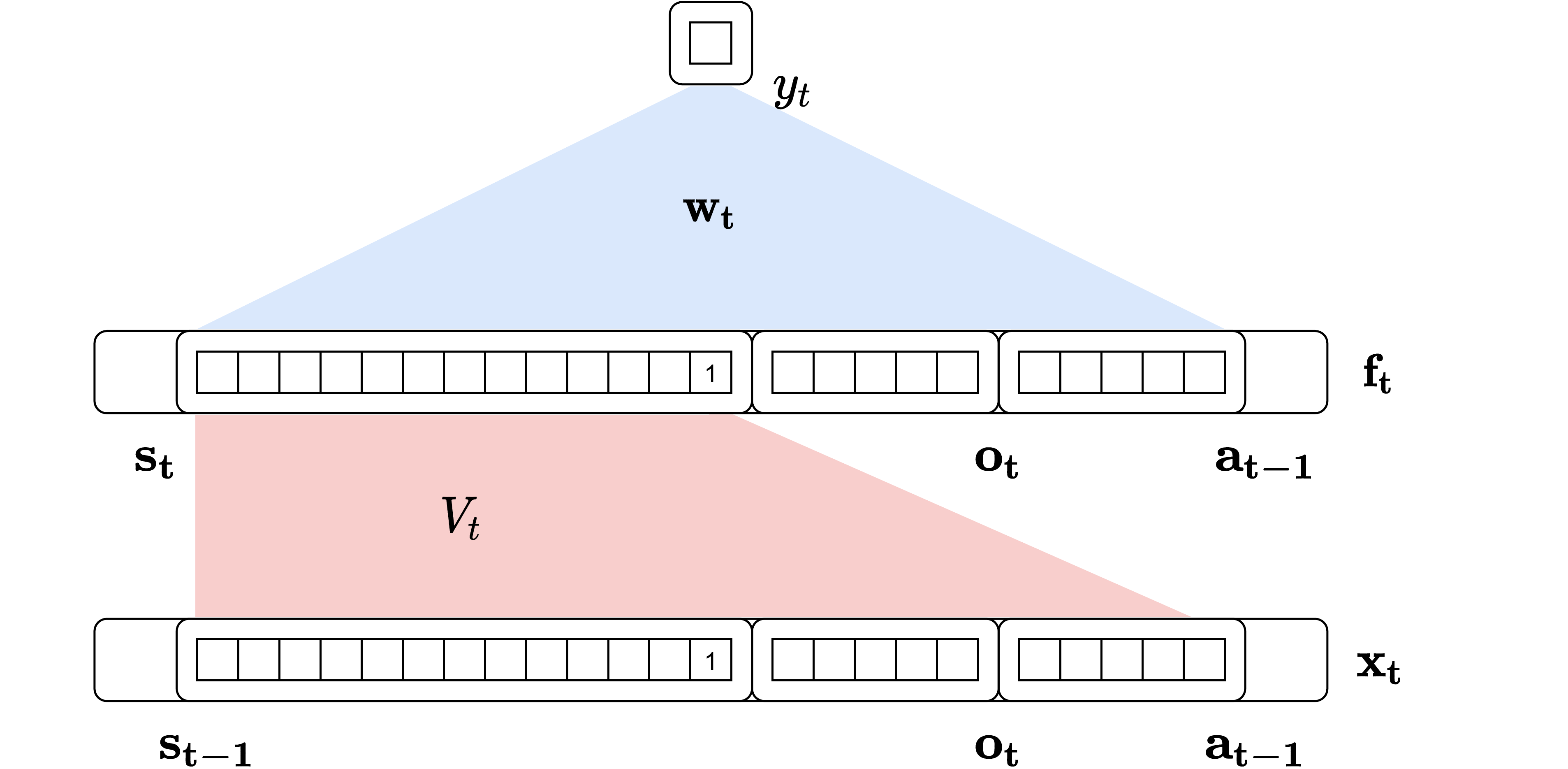}
\caption{The agent state $\mathbf{s}_t$ at time step $t$ is computed using the previous time step state $\mathbf{s}_{t-1}$ and the most recent observation $\mathbf{o}_t$ and action $\mathbf{a}_{t-1}$ using weight matrix $V$. The current agent state can then be used by the agent for prediction or control. For example, for the case of a single scalar prediction, the current state $\mathbf{s}_t$ and the most recent observation $\mathbf{o}_t$ and action $\mathbf{a}_{t-1}$ are mapped to the final prediction $y_t$ using weight vector $\mathbf{w}_t$. The weight matrix V is learned by a generate-and-test algorithm, and the weight vector $\mathbf{w}$ is learned by semi-gradient TD($\lambda$).}
\label{fig:agent_state}
\end{center}
\end{figure}

There can also be an always on bias bit in the agent in the agent state vector. An abstract view of agent state architecture is shown in Figure~\ref{fig:agent_state}. Note that the most recent observation $\mathbf{o}_t$ and action $\mathbf{a}_{t-1}$ are used to compute the current time step state $\mathbf{s}_t$ and are also directly connected to the final prediction $y_t$.  

A generate-and-test algorithm learns the weight matrix $V$. Generating a new feature $s_t^i$ corresponds to selecting which input signals to connect to in $\mathbf{x}_t$ and its connection weight in the weight matrix $V$. When we remove a feature, we delete its connections to the $\mathbf{x}_t$. In the case of learning an online multi-step prediction task, we use semi-gradient TD($\lambda$) \citep{Sutton1988TD}. Semi-gradient TD($\lambda$) is computationally efficient and can learn predictions online \citep{Modayil2014Nexting}. We use semi-gradient TD($\lambda$) to learn the weight vector $w$ to predict the future value of a cumulant $c$ \citep{Sutton2011Horde}---in the case of classical conditioning problem we can consider the US as the cumulant \citep{Sutton1990TDmodel}. Since the agent state is learned separately by a generate-and-test algorithm, we can use linear semi-gradient TD($\lambda$) to update the weight vector $\mathbf{w}$
\begin{align}
\label{eq:tde}
{\mathbf z}_t = \gamma\lambda {\mathbf z}_{t-1} + {\mathbf s}_t
\end{align}
\begin{align}
\label{eq:td}
{\mathbf w}_{t+1} = {\mathbf w}_{t}+\alpha(c_{t+1}+\gamma{\mathbf f}_{t+1}^\intercal{\mathbf w}_t - {\mathbf f}_{t}^\intercal{\mathbf w}_t){\mathbf z}_t
\end{align}
where $\alpha \in (0,1]$ is the step size and $\gamma \in [0,1)$ is the discount factor that determines the horizon of the prediction of the cumulant. $\mathbf{z}_t \in \mathbb{R}^{m+n+d}$ is the eligibility trace, and $\lambda \in [0,1]$ is the decay of the eligibility trace. 




\section{Deep Trace Generate-and-Test for Trace Conditioning}
\label{sec:dtrace-generator}
The first problem we focus on is how to generate features that enable the agent to associate temporally distant events. Using the trace conditioning problem, we can focus on this problem in isolation. The trace conditioning problem considers the problem of uncontrolled multi-step prediction in which two stimuli that have no prior association are presented to the animal in a particular order. First, a CS is presented to the animal, followed by an US. The animal responds to the US by generating an unconditioned response (UR). After several trials, the animal generates a conditioned response (CR) when presented with the CS. The UR happens before the arrival of the US, which suggests that the animal associated the CS with the US. For example, a dog would naturally salivate (UR) in the presence of food (US).  Suppose the dog receives a previously unassociated stimulus such as a tone (CS) before the arrival of the food. After enough pairings, the dog will start salivating (CR) after hearing the tone and before the arrival of the food. There can be a gap between the offset of the CS and the onset of the US. This gap is called the trace interval.  There is no immediate relevant observation available during the trace interval; thus, the agent should make features that fill the gap to predict the US accurately. 

The TD model of classical conditioning can make predictions similar to predictions observed by animal experiments \citep{Sutton1990TDmodel}. The TD model of classical conditioning uses TD($\lambda$) to learn predictions \citep{Sutton1988TD}; however, to make accurate predictions, the state should have features representing the trace interval \citep{Ludvig2012EvaluatingTDmodel}. To make features representing the trace interval, we introduce our first generator, the \textit{deep trace generator}. The deep trace generator makes features that enable the agent to remember helpful information from the past to predict the future. We call these features \textit{deep trace features}. Deep trace features are \textit{traces} from either observation signals or other features---including other deep traces.

The deep trace feature $s^{i}$ traces $x^j$ (either another feature or an observation signal) by connecting to itself at the previous time step with the weight $\psi \in (0,1)$ and to $x^j$ with the weight $1-\psi$. The deep trace feature $s^i$ gets computed at every time step as follows:
\begin{align}
\label{eq:dtrace}
s^{i}_{t} = \psi s^{i}_{t-1} + (1-\psi) x^{j}_{t}
\end{align}
in which $\psi$ is the \textit{decay rate} and $x^j$ the \textit{source} of the deep trace feature $s^i$. 

For example, let $s^a$ to be a deep trace feature of the observation signal $o^j$ with a decay rate of 0.9. At time step 2, $o_j^2$ becomes 1 for one time step. Assuming $s^a_{1}=0$ (since $o^j_1=0$), we can calculate $s^a_1$ using Equation~\ref{eq:dtrace}
\[s^{a}_{1} = 0.9 s^{a}_{0} + (1-0.9) o^{j}_{2} = 0.9(0) + 0.1(1) = 0.1.\]
At time step $2$, the observation $o^j_2$ is no longer present ($o_j^{2}=0)$; however, since $s^{i}_{1}=0.1$, $s^{i}_{2}$ becomes 0.09. Even if $o^j$ remains inactive for more time steps, the deep trace feature $s^a$ will have non-zero activity for many more time steps. The trace decay rate $\psi$ controls how rapidly the deep trace fades away. The self-connection weight $\psi$ and source connection weight $(1-\psi)$ ensure the value of the deep trace remains between 0 and 1---assuming the source of the deep trace is binary or is between 0 and 1. 

\begin{figure}[ht]
    \centering
    \begin{subfigure}[t]{0.5\textwidth}
        \includegraphics[width=\textwidth]{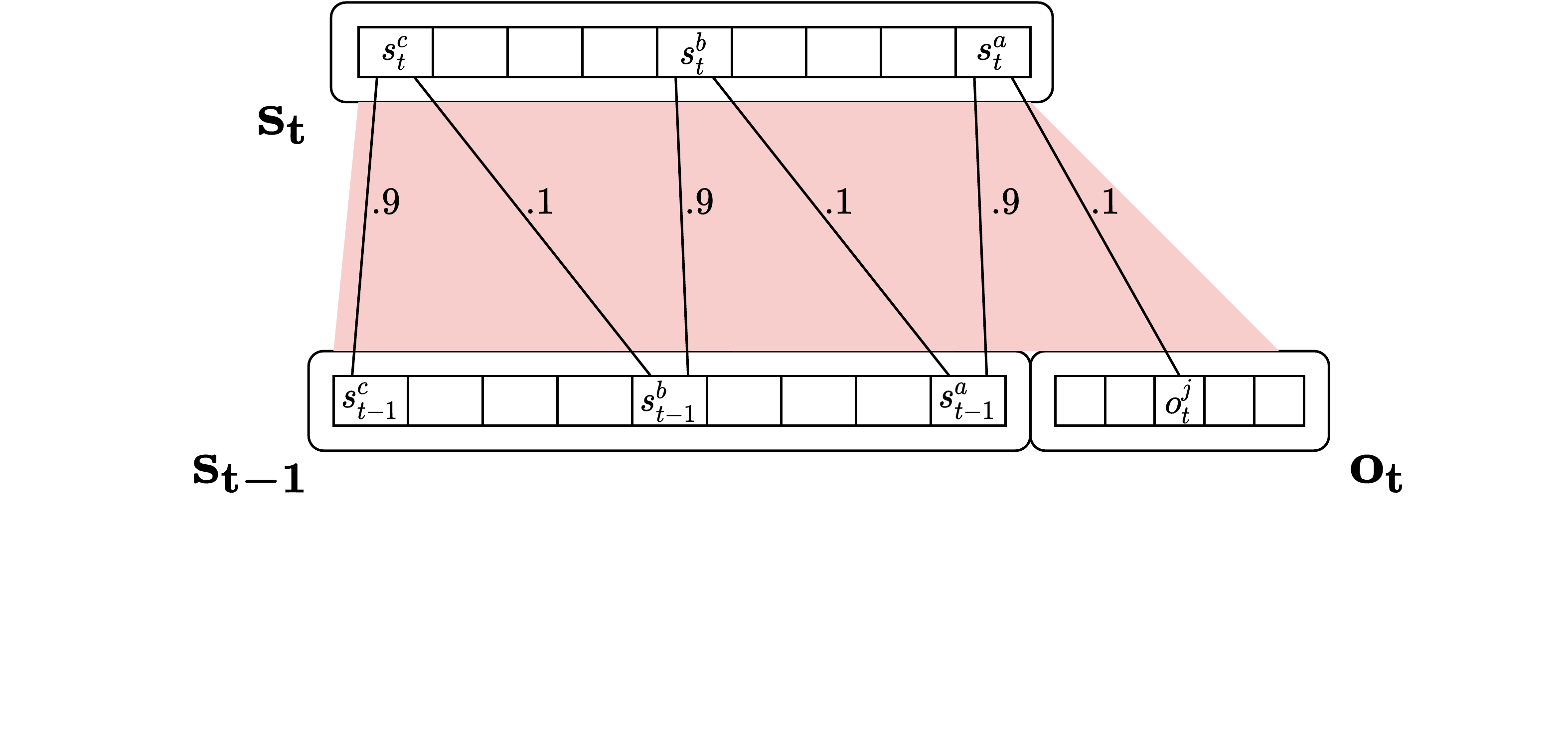}
    \end{subfigure}
    \begin{subfigure}[t]{0.45\textwidth}
        \includegraphics[width=\textwidth]{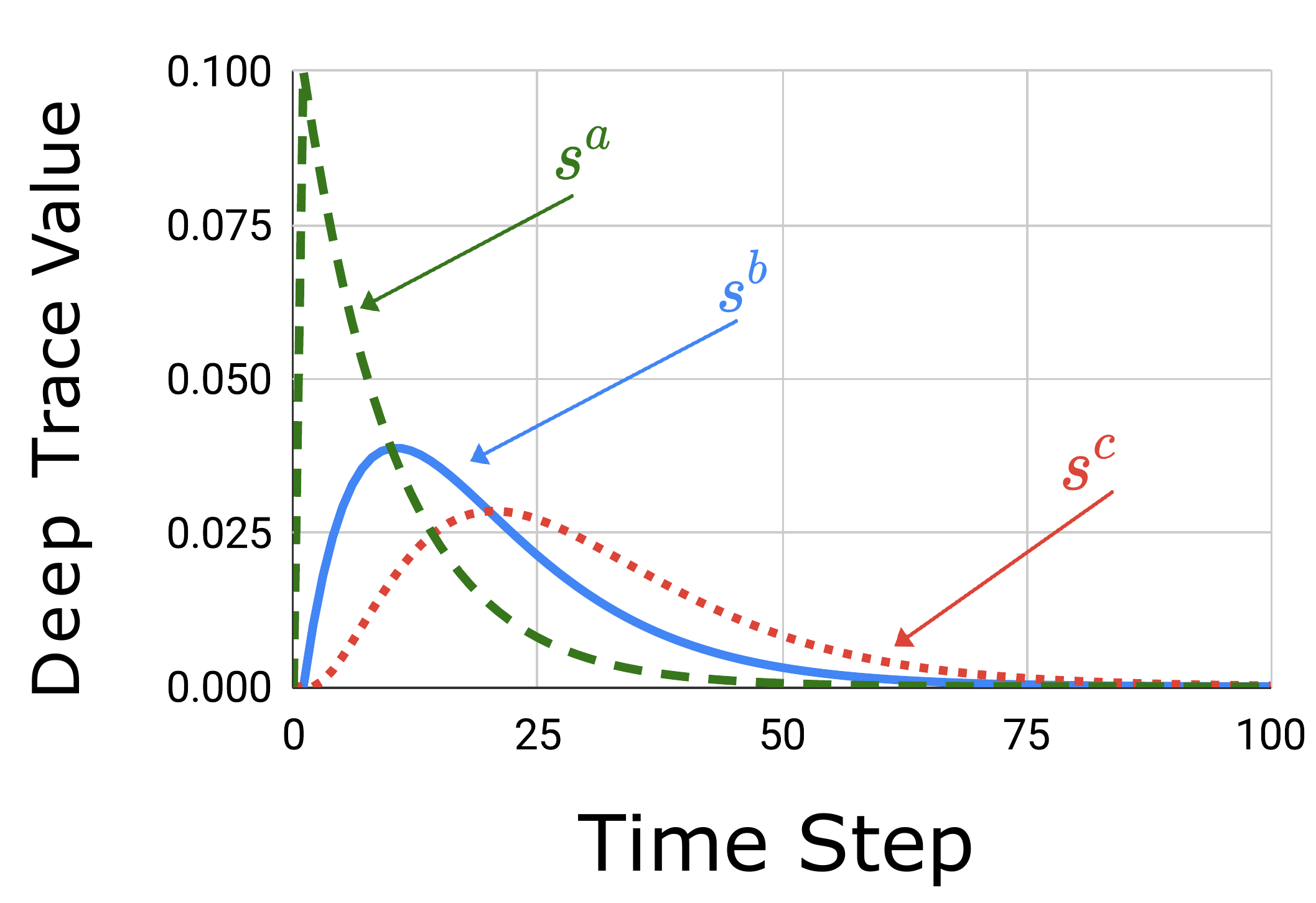}
    \end{subfigure}

    \caption{The abstraction of deep trace features $s^a$, $s^b$, and $s^c$ that trace $o^j$, $s^a$, and $s^b$, respectively (right figure). When the observation signal $o^j_2$ gets activated at time step 2 for one time step, deep trace feature $s^a$ quickly jumps up and starts to fade away. Deep trace feature $s^b$ and $s^c$ indirectly trace the observation signal $o^j$ and are slower to respond. The figure on the right shows the level of deep trace feature $s^a$, $s^b$, and $s^c$ over time. These deep trace features provide a rich memory of the observation signal $o^j$ that can allow the agent to fill the trace interval gap and make accurate predictions.}
    \label{fig:threedtrace}
\end{figure}

A deep trace feature can be the source of another deep trace feature . Let us extend our example by adding two more deep trace features. Deep trace features $s^b$ and $s^c$ trace deep trace feature $s^a$ and $s^b$, respectively---both of these new deep trace features have a decay rate of 0.9. Consequently, deep trace features $s^b$ and $s^c$ are indirect traces of the observation $o^j$.  Figure~\ref{fig:threedtrace} shows the value of deep trace features $s^a$, $s^b$, and $s^c$ over time. Since deep trace $a$ is directly connected to the observation $o^j$, it quickly jumps up and fades away faster than deep trace features $s^b$ and $s^c$. Deep trace features $s^b$ and $s^c$ are slower in growth but last longer.  Multiple traces of an observation signal (both indirect and direct) let the agent have a longer-lasting memory of that observation signal. In the trace conditioning problem, deep trace features enables the agent to fill the trace interval gap and remember the CS.

The deep trace generator makes a new deep trace feature by choosing the source and the decay rate. The generator chooses the decay rate randomly (between 0 and 1). The generator also selects the source randomly from the $\mathbf{x}_t$. However, the probability of each $x_t^i$ to be selected is proportional to its outgoing weight magnitude. In other words, $x_t^i$ with a larger outgoing weight magnitude $w_t^i$ is more likely to be selected as the source for a new deep trace feature.

The maximum number of features is fixed as the computation and memory cost should remain constant. To generate new features, the tester should remove the least useful features. At each time step, the tester partitions the features by half based on the exponential moving average of their weight magnitude. A certain number of features in the bottom half of the partitioning are subjected to deletion to make space for the generator to make more features. However, if a feature is the source of another deep trace, the tester refrains from removing it. This protection happens since a feature that ranks low in the partitioning can be a source for a useful feature.The tester protects a newly generated feature by initializing its moving average of the weight magnitude as the median of all moving averages of the weight magnitude. 

To show the effectiveness of the deep trace generator, we experiment on the trace conditioning problem \citep{Rafiee21Testbed}. Trace conditioning problem consists of several trials. Each trial starts with the CS lasting for 4 time steps followed by the US lasting for 2 time steps. The time from the onset of the CS and the onset of the US is called inter-stimulus interval (ISI). The time onset of the US and beginning of the next trial is the inter-trial interval (ITI) which we uniformly sampled from (80,120). In addition to the CS and the US, other uninformative stimuli happen randomly during each trial. These uninformative stimuli are called \textit{distractors} and provide no information about the CS or the US. In our experiments, there are 10 distractors that occur with a Poisson distribution and the rate of $\frac{1}{10}$,$\frac{1}{20}$, ..., $\frac{1}{100}$, respectively and lasted for 4 time steps. The discount factor $\gamma$ is set to $1-\frac{1}{ISI}$. To evaluate the performance of the agent we use the Mean Squared Return Error (MSRE) over bins of 1000 time steps. The Squared Return Error (SRE) at time t is calculated by $(y_t-G_t)^2$ in which $G_t = \sum_{k=0}^{\infty} \gamma^k c_{t+k+1}$ is the return.

\begin{figure}[ht]
    \centering
    \begin{subfigure}[t]{0.3\textwidth}
        \includegraphics[width=\textwidth]{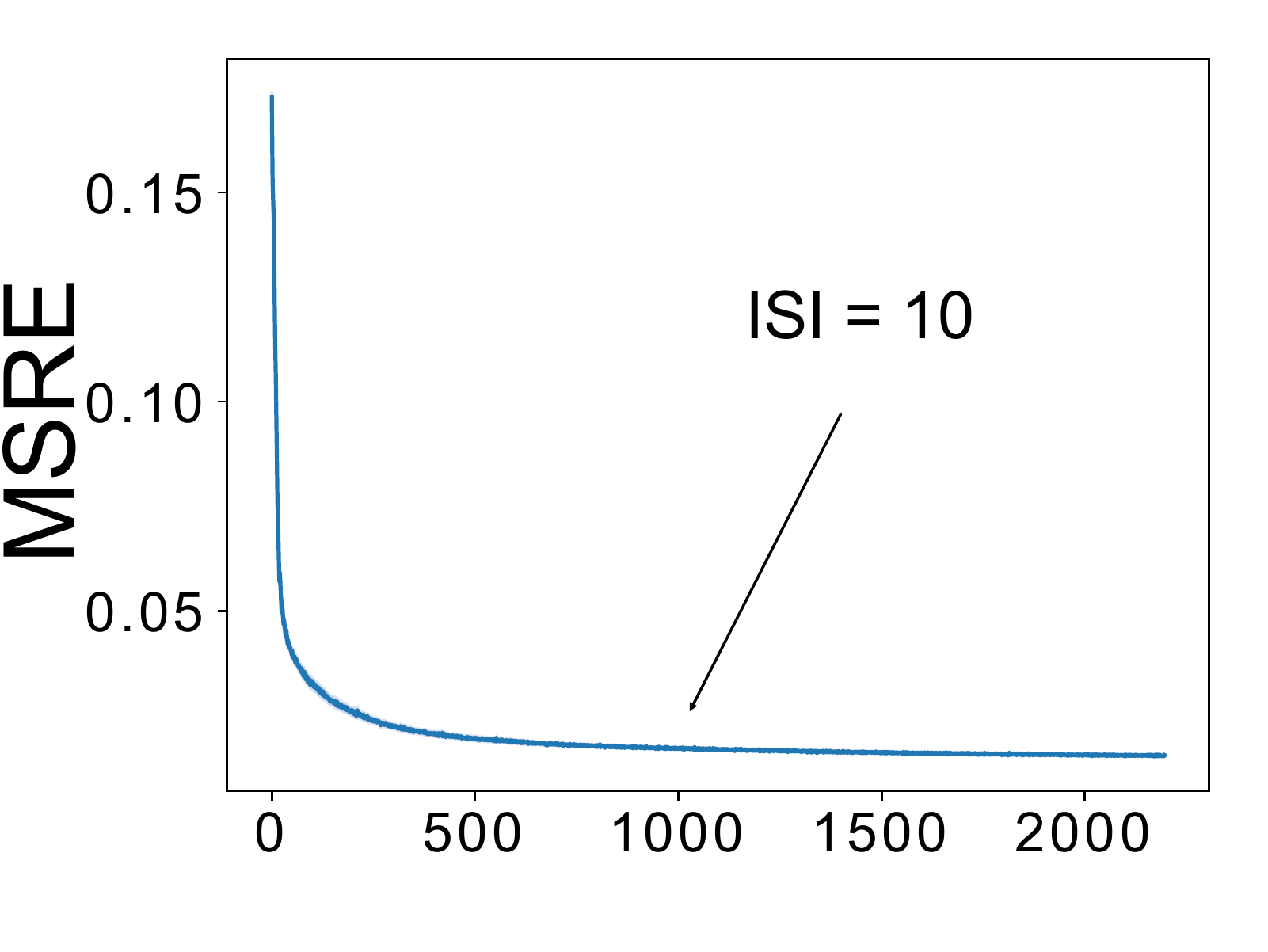}
    \end{subfigure}
    \begin{subfigure}[t]{0.3\textwidth}
        \includegraphics[width=\textwidth]{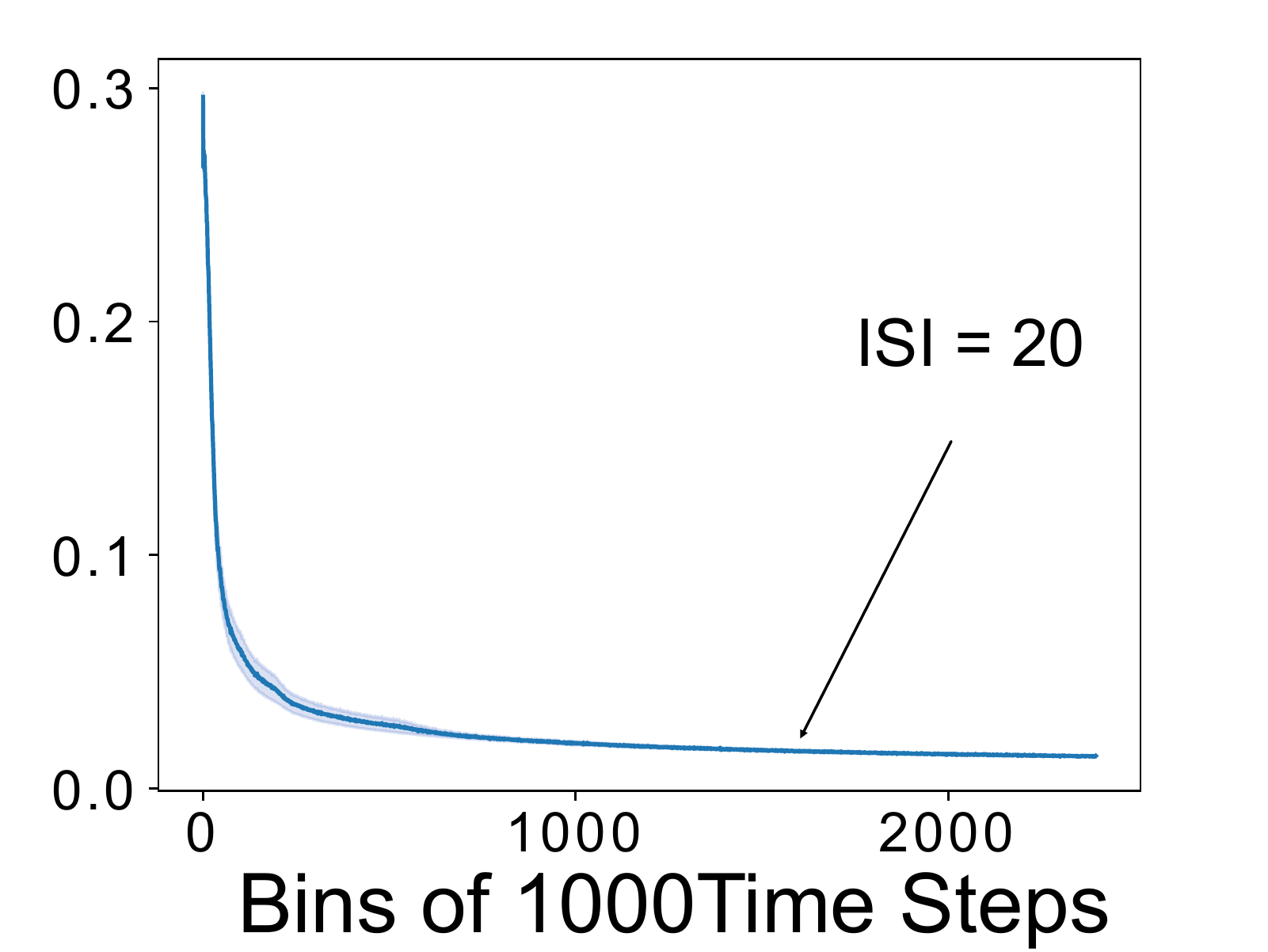}
    \end{subfigure}
        \begin{subfigure}[t]{0.3\textwidth}
        \includegraphics[width=\textwidth]{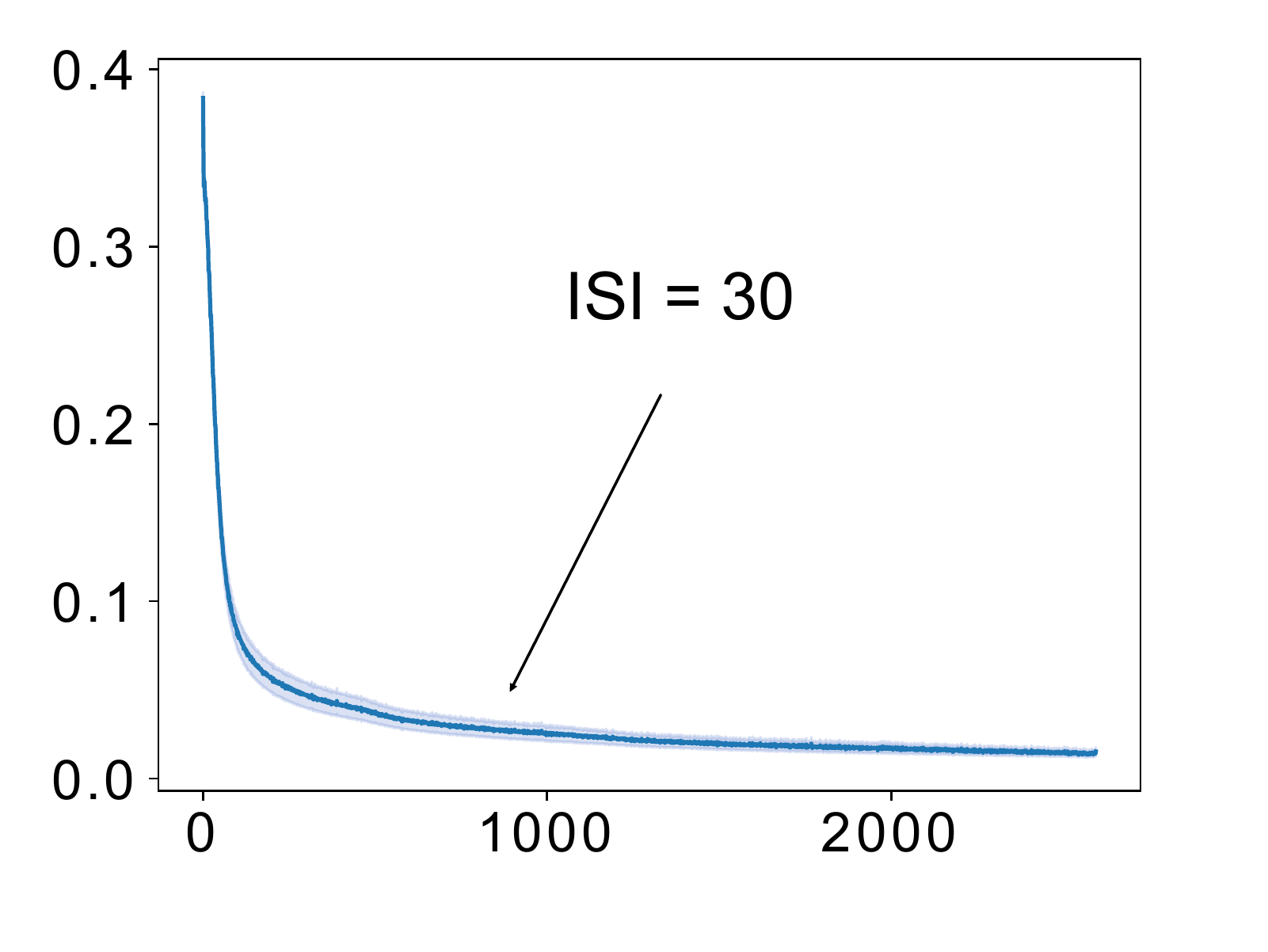}
    \end{subfigure}
    \caption{The MSRE over bins of 1000 time steps for the deep trace generator on the trace conditioning problem with varying ISI 10, 20 , and 30 averaged the MSRE over 30 runs. The shaded area is the standard error. The agent can make accurate predictions using the features learned by the deep trace generator.}
    \label{fig:TC_rmse}
\end{figure}

\begin{figure}[ht]
    \centering
    \begin{subfigure}[t]{0.3\textwidth}
        \includegraphics[width=\textwidth]{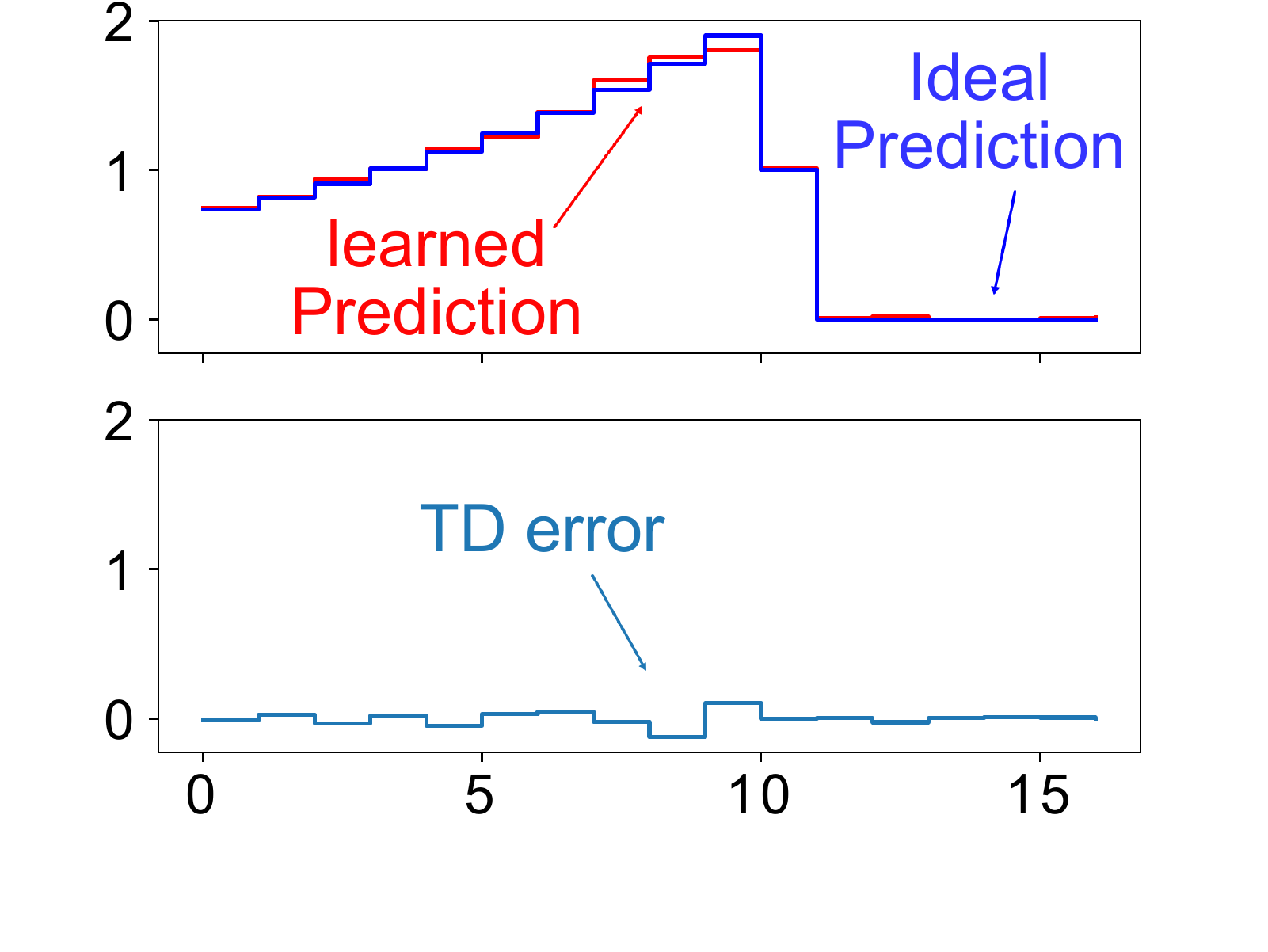}
    \end{subfigure}
    \begin{subfigure}[t]{0.3\textwidth}
        \includegraphics[width=\textwidth]{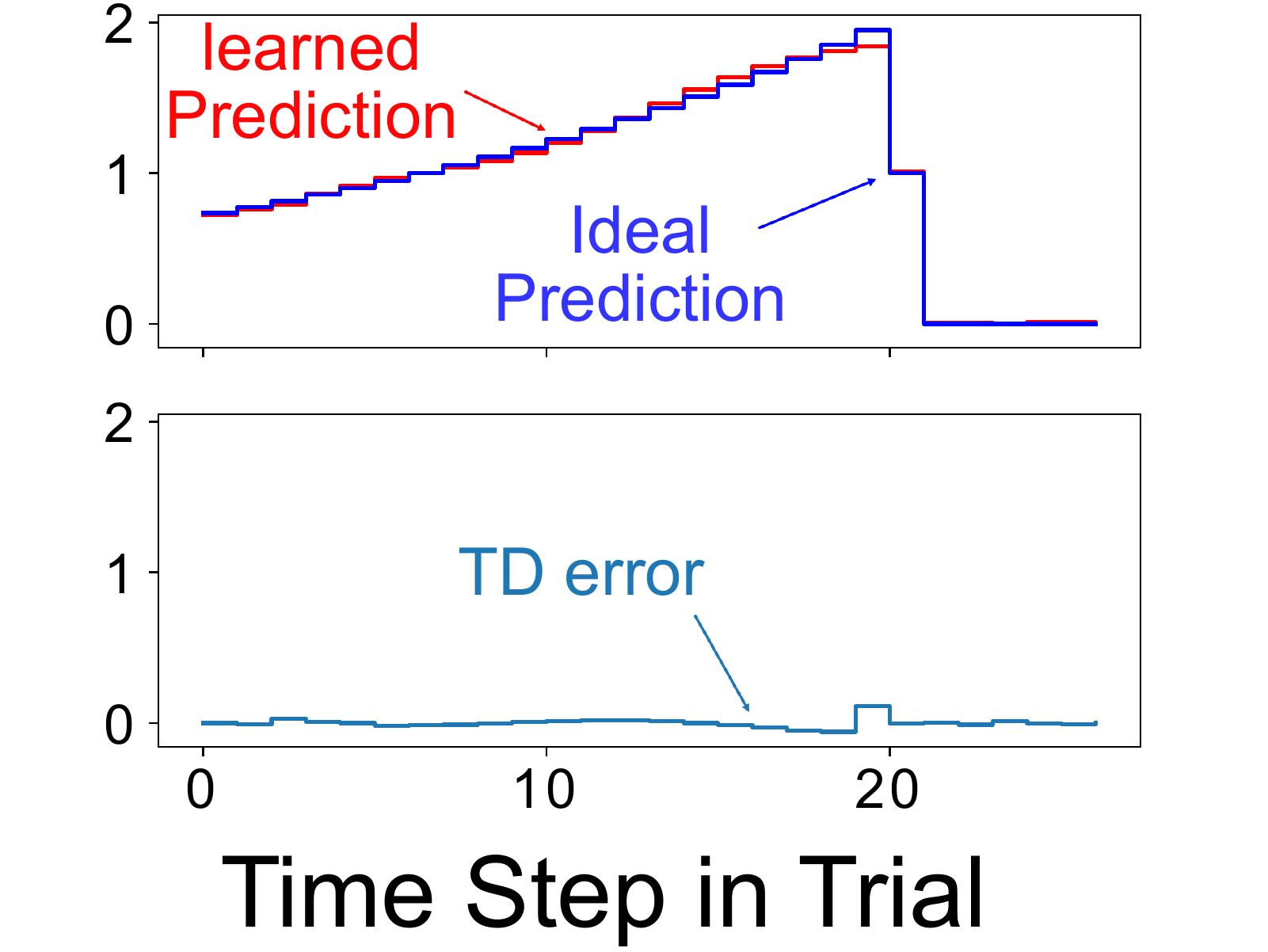}
    \end{subfigure}
        \begin{subfigure}[t]{0.3\textwidth}
        \includegraphics[width=\textwidth]{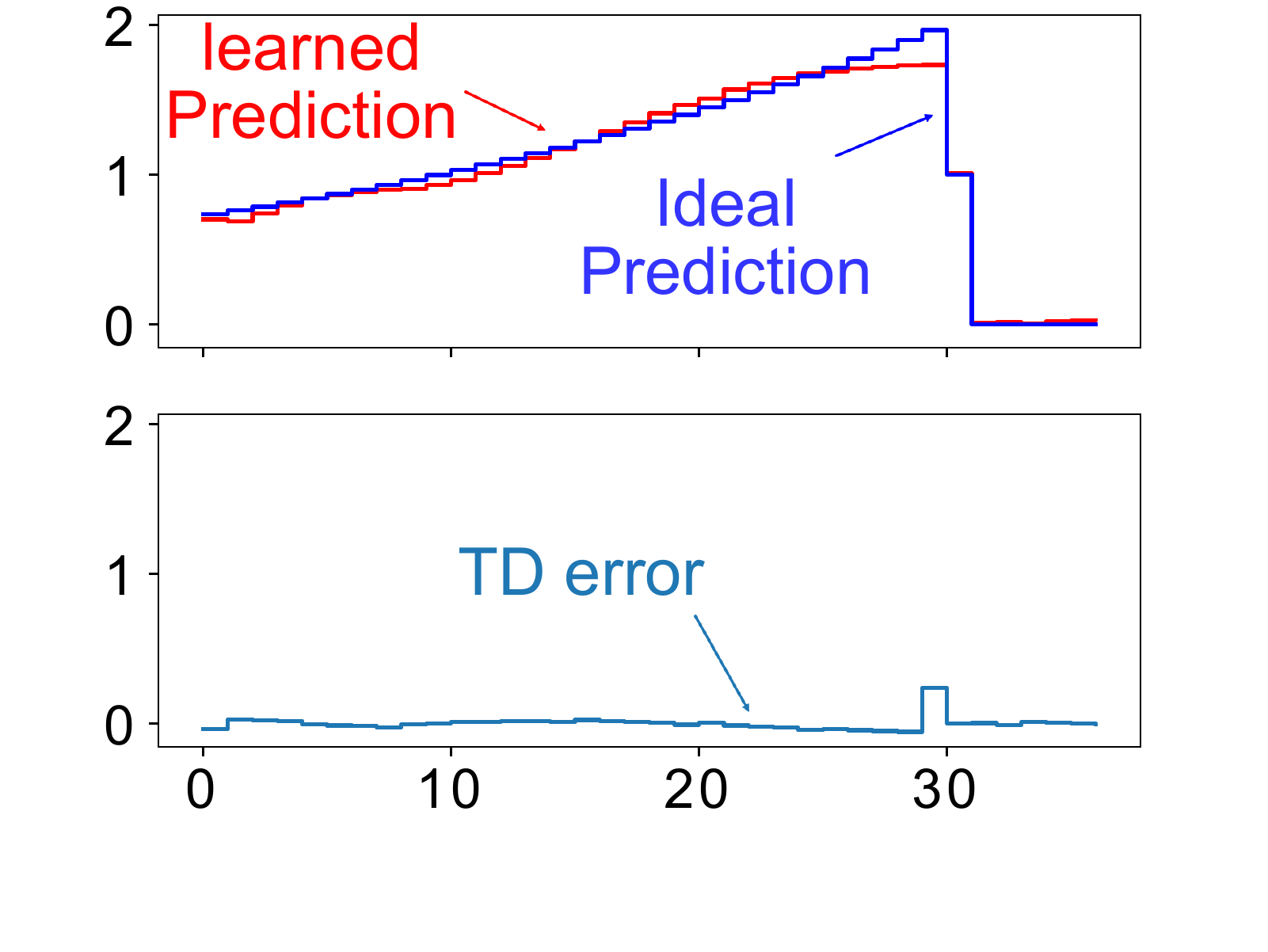}
    \end{subfigure}
    \caption{Prediction made using the features generated by the deep trace generator after more than 2 million time steps (red line) compared to the ideal prediction (blue line) which is the return. The bottom plots show the TD error. The agent's prediction closely matches the ideal prediction, which shows the agent could effectively fill the trace interval gap.}
    \label{fig:TC_trial}
\end{figure}

We evaluate the effectiveness of the deep trace generator for enabling the agent to remember distant stimuli by running our experiments for ISI of 10, 20, and 30. Each experiment consists of 20000 trials for each ISI value, which is more than 20 million time steps. We use semi-gradient TD($\lambda$) with $\lambda=0.9$ and we use step-size adaptation with initial step-size of $\alpha=0.01$ and meta step-size $\theta=0.01$ \citep{sutton1992idbd,thill2015tidbd}. The maximum number of features is set to 100, 200, and 300 for ISI 10, 20, and 30, respectively. The maximum number of features to add ($g_d$) and remove ($r_d$) at each time step is set to 2. Figure~\ref{fig:TC_rmse} shows the MSRE over bins of 1000 time steps. We average MSRE over 30 runs, and the shaded area is the standard error. Figure~\ref{fig:TC_trial} shows the learned predictions by the agent and compares them to ideal predictions based on the return. The agent is able to make accurate predictions using the features generated by the deep trace generator.

\section{The Imprinting Generator for Trace Patterning}
\label{sec:impc_generator}
In the trace conditioning problem, the agent only needs to remember a single CS to predict the US. Although there are distractors, they provide no information, and the agent can ignore them and focus only on the CS and the US. In trace patterning, there are multiple CSs, and only a specific configuration of active and inactive CSs triggers the arrival of the US. For instance, the dog would receive food only if the tone is present and the light is absent. Without considering the non-linear configurations of the CSs, each CS would not be a useful predictor of the US. The deep trace generator could only generate traces of individual features or observation signals. We need to generate features that respond to configurations in the observation signals.

We propose the \textit{imprinting generator} for generating features that respond to a particular configuration in the observation signals. An \textit{imprinting feature} $s^i$ is connected to observation signal $o^j$ with a weight of +1 if $o^j$ should be active and -1 if $o^j$ should be inactive in the configuration. Note that not all observation signals need to be connected to $s^i$. The imprinting feature $s^i$ is a non-linear map of the observations that are connected and is computed using Linear Threshold Unit (LTU) \citep{sutton1993err}. The imprinting feature $s^i$ is computed as follows:
\[   
s_t^i = 
     \begin{cases}
       \text{1} &\quad\sum_{j=1}^{m} V_t^{i,j}o_{t}^{j} \geq \sum_{j=1}^{m} V_t^{i,j}  \\
       \text{0} &\quad\text{otherwise} \\
     \end{cases}
\]

\begin{figure}[ht]
\begin{center}
\includegraphics[width=10cm]{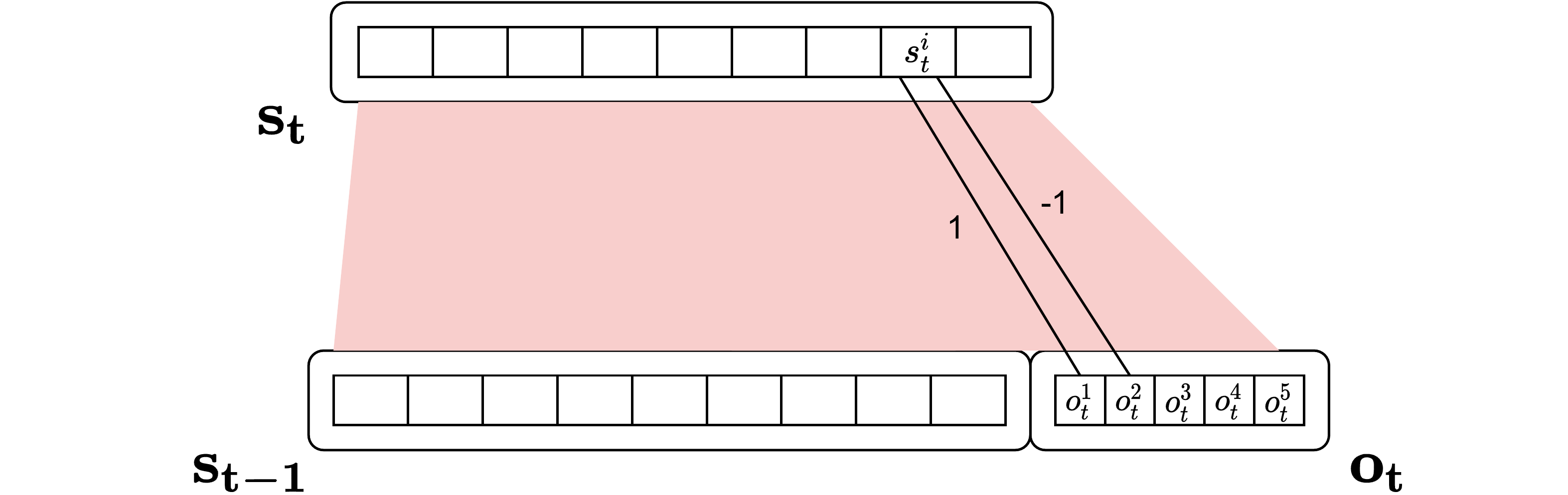}
\caption{Imprinting feature $s^i$ is connected to the observation signal $o^1$ with a weight of +1 and the observation signal $o^2$ with a weight of -2. The imprinting feature $s^i_t$ would become activated (1) if the $o_t^1=1$ and $o_t^2=0$. The imprinting feature $s_i$ represents a non-linear configuration of its connected observation signals.}
\label{fig:configuration}
\end{center}
\end{figure}

Figure~\ref{fig:configuration} shows an example in which the imprinting feature $s^i_t$ is connected to the observation signal $o^1_t$ and $o^2_t$ with a weight of +1 and -1, respectively. The imprinting generator needs to decide which observation signals to connect to. A simple solution would be selecting the observation signals randomly, but the space of possible imprinting features is $3^m$ (connecting with +1 or -1 or no connection), which can slow down learning significantly. Since the observation vector $\mathbf{o}_t$ has direct weights to the prediction $y_t$, we use those weights to make the observation signals with a larger weight magnitude more likely to participate in imprinting features. Observation $o^i_t$ participates in the creation of a imprinting feature at time step $t$ if 
\begin{align}
\label{eq:config}
\frac{|w_t^{i+n}|}{\sum_{j=n+1}^{m+n}|w_t^j|} \geq \frac{1}{m} + \epsilon  
\end{align}
where $\epsilon \sim \mathcal{N}(0,\,\frac{1}{m})\,$ is small random noise to give observation signals with small weight a chance to be selected. The connection weight is +1 if the observation $o^i_t$ is active and -1 if the observation is inactive at time $t$. There is also the question of when we generate imprinting features. The imprinting generator monitors the observation signals. Suppose there is a non-zero activity in the observation signals. In that case, the imprinting generator makes $g_c$ new imprinting features (if there is capacity in the network) and add these features if they are new---not to add duplicate features to the network. The imprinting generator works alongside the deep trace generator. Together, these generators would make non-linear combinations of observations and remembers them to make temporally distant associations.

The tester is the same as our tester for the trace conditioning problem. However, the maximum number of features in the network is divided between the two generators. Otherwise, the deep trace generator would end up using almost all of the capacity. The tester also applies to the deep traces and imprinting features separately. Thus, each feature would be compared only to the features of the same type.  Algorithm~\ref{alg:complete_algorithm} demonstrate the details of our proposed generators and tester. 

\begin{algorithm}[!h]
\textbf{Initialize:} Set the state-update function $u$ with no initial features and consider the agent state $\mathbf{s}_{-1} \in \mathbb{R}^n$ as zeros\\
\textbf{Initialize:} Set weight vector $\mathbf{w}\in \mathbb{R}^{n+m}$ and eligibility trace vector $\mathbf{z}\in \mathbb{R}^{n+m}$ as zeros\\
\textbf{Initialize:} Set hyper-parameters $\alpha$, $\theta$, $\lambda$, $c_d$, $g_d$, $r_d$, $\mu$, $p_d$, $c_i$, $g_i$, $r_i$, $\mu$, and $p_i$\\ 
\vspace{2mm}

\For{each observation {$\mathbf{o}_{t}$} and {$\mathbf{US}_t \in \mathbb{R}$}}{
\If{there is non-zero activity in $\mathbf{o}_t$ and $c_i$ is not reached}{
Generate $g_i$ number of imprinting features \\
\For{each generated feature $i$}{ 
    Select the observations using Equation~\ref{eq:config}\\
    Add the feature $i$ if it is a new feature---not to add duplicate features.
}
}
Compute the current state: $\mathbf{s}_t = u(\mathbf{s}_{t-1},\mathbf{o}_{t})$ \\
Compute the prediction: $y_t = \mathbf{w}_{t}^T \mathbf{s}_{t}$\\
$\delta_t = US_{t} + \gamma y_t - y_{t-1}$\\
$\mathbf{z}_t = \gamma \lambda \mathbf{z}_{t-1} + \nabla_{\mathbf{w}}v_t$ \\
$\mathbf{w}_{t+1} = \mathbf{w}_{t} + \alpha \delta_t \mathbf{z}_t$\\
\If{$c_d$ is not reached}{
Generate $g_d$ deep trace features\\
\For{each generated feature $i$}{ 
Set the decay rate $\psi$ randomly \\
Choose the source  $j$ randomly \\
Set $V^{i,j}$ to $1- \psi$\\
Set $V^{i,i}$ to $\psi$\\
}
}
\If{$c_d$ is reached}{
Remove $r_d$ features from the bottom $1 - p_d$ portion of the deep trace features---based on the weight magnitude---that are not a source for other features \\
\For{each removed feature $i$}{ 
Set the outgoing weight $w^i$ to 0\\
Set the corresponding eligibility trace $z^i$ to 0\\
}
}
\If{$c_i$ is reached}{
Remove $r_i$ features from the bottom $1 - p_i$ portion of the imprinting features---based on the weight magnitude---that are not a source for other features \\
\For{each removed feature $i$}{ 
Set the outgoing weight $w^i$ to 0\\
Set the corresponding eligibility trace $z^i$ to 0\\
}
}
}
\DontPrintSemicolon
\SetKwInOut{Input}{Input}
\SetKwInOut{Output}{Output}

\caption{Imprinting and Deep trace generate-and-test algorithm.}
\label{alg:complete_algorithm}
\end{algorithm}

We study the effectiveness of the imprinting generator on the trace patterning problem. In our setup, there are 6 CSs and 10 distractors. All CSs and distractors have a duration of 4 time step if they become active. A specific configuration of 3 active and 3 inactive CSs would result in the arrival of the US, which is referred to as the activation pattern. In our experiments, the activation pattern occurs in half of the trials. The distractors are also presented to the agent simultaneously with the CSs. Each distractor occurs independently with a probability of 0.5. If the activation pattern is present during a trial, the US arrives after 10 time steps (ISI=10) and remains active for 2 time steps. The discount factor $\gamma$ is set to 0.9. We evaluate the performance of the agent using MSRE over bins of 1000 time steps. All the hyper-parameters are the same as the trace patterning experiments, except the maximum number of deep traces is set to 200, and the maximum number of imprinting features is set to 60. The maximum number of imprinting features to add ($g_c$) and remove ($r_d$) at each time step is set to 2. 

\begin{figure}[ht]
    \centering
    \begin{subfigure}[t]{0.45\textwidth}
        \includegraphics[width=\textwidth]{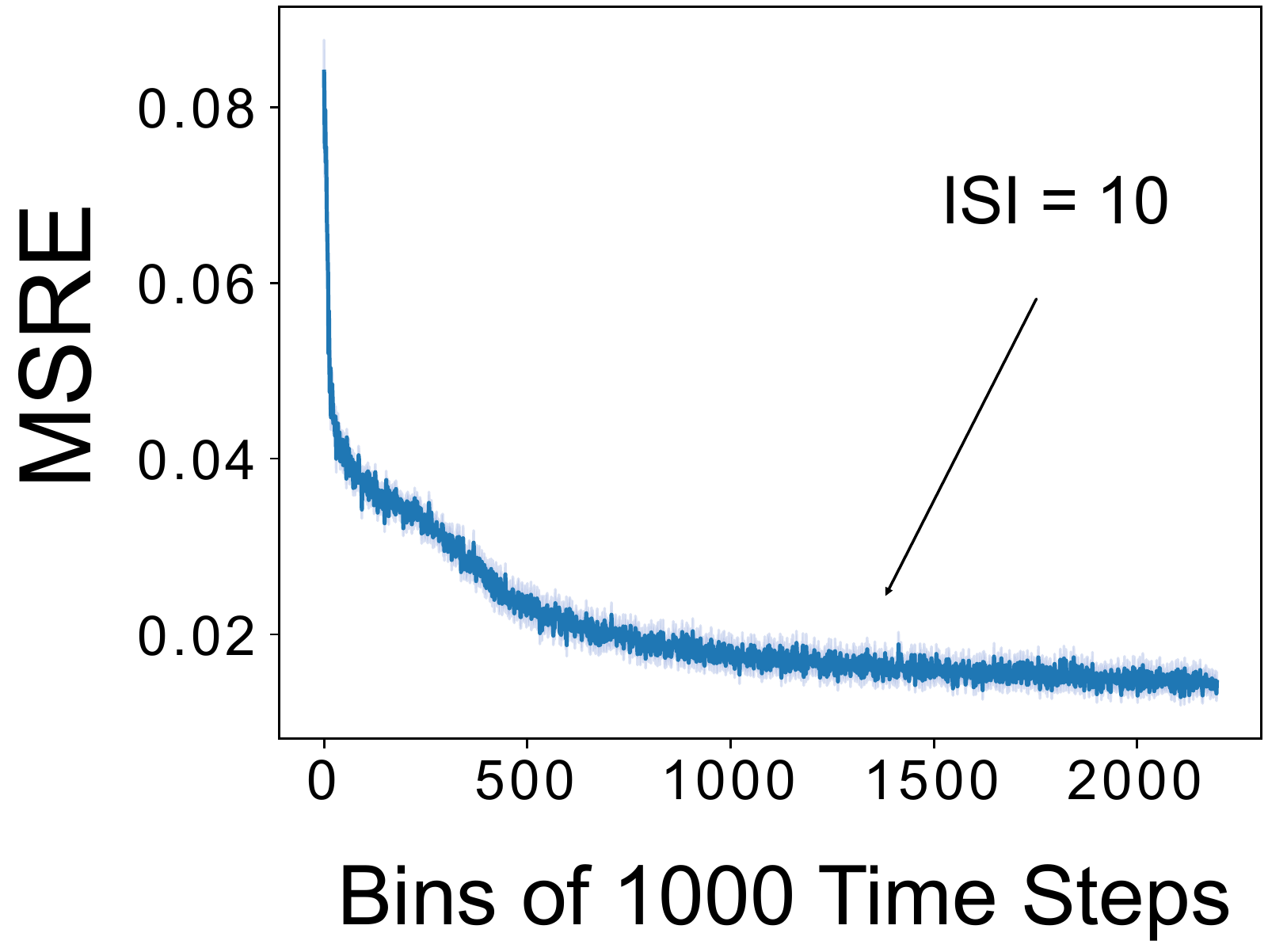}
    \end{subfigure}
    \begin{subfigure}[t]{0.45\textwidth}
        \includegraphics[width=\textwidth]{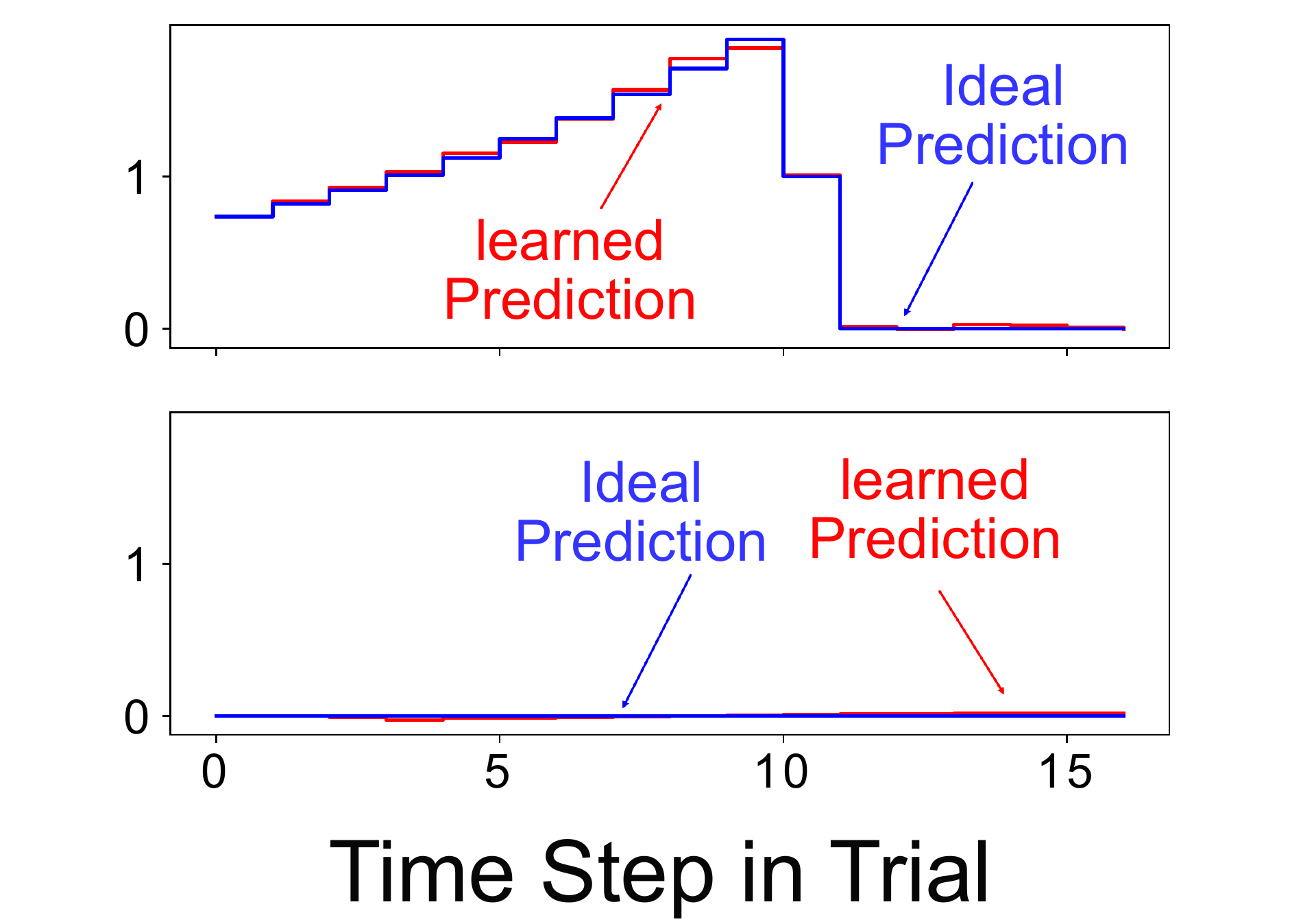}
    \end{subfigure}
    \caption{The MSRE over bins of 1000 time steps for trace patterning problem averaged over 30 runs and the shaded area is the standard error (left figure). The prediction learned using the features generated by the imprinting generator and the deep trace generator after more than 2 million time steps (red line) compared to the ideal prediction (blue line) based the return. The agent can accurately predict the arrival of the US (top right) and its absence (bottom right), suggesting that the agent represent the useful non-linear configuration of observation signals and fill the trace interval gap. }
    \label{fig:TP_rmse_trial}
\end{figure}

The focus of this experiment is to show the effectiveness of the imprinting generator in finding the correct configuration. There are $3^{16}$ possible configurations (6 CSs and 10 distractors). We run the experiment for 20000 trials which is about 20 million time steps. We report the results for this experiment in Figure~\ref{fig:TP_rmse_trial}. Figure~\ref{fig:TP_rmse_trial} compares the agents learned prediction and ideal prediction based on the return. The agent can accurately predict the arrival of the US in the case that the activation pattern occurs. The agent also makes accurate predictions when the US would not be presented.

\section{Related Studies}
\label{sec:related}
Employing search for finding representation has been studied in the supervised learning setting. \cite{sutton1993err} introduce the random representation for online learning, and \cite{Mahmood13GenTest} introduce generate-and-test by searching for random features that improve the performance of the base system. Later on, \cite{Dohare2021CBP} show that training neural networks using stochastic gradient descent is not ideal for the continual learning setting and suggest that when the initial randomness in the weights is lost, the performance drastically degrades. This issue is mitigated by performing generate-and-test alongside Backpropagation. The Cascade correlation learning architecture introduced by \cite{fahlman1990cascade} constructively add layers to construct a neural network. Studies mentioned above focus on feed-forward networks, and they are not directly applicable to learning the agent state for a reinforcement learning agent.

Learning the state for a reinforcement learning agent using the data stream of experience is a challenging problem. Recurrent neural networks are often used to learn the agent state. The difficulty is how to train these recurrent neural networks. Backpropagation through time (BPTT) \citep{elman1990finding} is a solution based on stochastic gradient descent that updates the weight of recurrent neural network to minimize the error. BPTT requires storing all previous network activations to unroll and update the weight, making it prohibitively expensive for online learning. Truncated BPTT (TBPTT) \citep{williams1990tbptt} only store the past $t$ number of activations which makes the complexity of training constant, but it is still expensive. The truncation parameter $t$ is where we decide not to consider further dependencies. Longer truncation enables the network to consider temporal association further in the past at the expense of memory and computation. RTRL \citep{williams1989learning} is another alternative to BPTT that allows for online training, but the cubic computation makes it intractable for larger networks. Approximating the gradient for RTRL makes it more computationally feasible; though, these approximations introduce parameters that their choice influences how far back in time we can make associations \citep{Menick2020SparseRTRL,tallec2017uoro}. Gated RNNs \citep{chung2014GRU,hochreiter1997LSTM} modify the architecture of the RNNs to mitigate problems such as vanishing gradients that make long temporal associations difficult \citep{hochreiter2001gradient}; however, These architectures still need training algorithms such as TBPTT or RTRL, which are not suited for online learning. 

Predictive representation methods can learn the state by answering predictive questions. TD networks learn the agent state by learning a network of predictions using TD methods \citep{sutton2005tdnet,tanner2005tdlnet}. OTD network \citep{rafols2006otdnet} is an extension of the TD network that includes temporal abstracted predictive questions also known as options \citep{sutton1999options}. More recently, General Value Function Networks \citep{schlegel2021gvfn} restrict the hidden state of a recurrent neural network to be predictive questions. The challenging part of the predictive representation methods is discovering what predictive questions the agent should consider. 

\section{Discussion}
\label{sec:disc}
Learning the agent state online is an essential step towards more general reinforcement learning agents. This work demonstrates how to learn the agent state by searching for features that improve the agent's performance on online partially observable multi-step prediction tasks. We focused on two benchmark problems introduced by \cite{Rafiee21Testbed}. First, using the trace conditioning problem, we focus on the agent's ability to predict an upcoming stimulus based on a temporally distant cue. For this problem, we introduce the deep trace generator. We show the effectiveness of the deep trace generator and tester on three instances of trace conditioning problems by varying the ISI. Second, the trace patterning problem is used as an extension of the trace conditioning problem to the case where a non-linear configuration of stimuli results in the arrival of the US. For this problem, we propose the imprinting generator. The imprinting generator makes features that only get activated when a particular configuration in the observation signals occurs. We show that the deep trace generator and imprinting generator can learn useful non-linear configurations of observation signals and remember them for accurate predictions of the US.

The agent state is used in the agent's policy, value functions, and model of the environment. We only studied learning the agent state when there is only a single user for it. Our simple tester used the weight magnitude for the prediction as an indicator of the usefulness of a feature. When there are multiple users for the agent state, it is unclear which features to preserve and which ones to delete. Generate-and-test is a promising approach towards learning the agent state, and future research should investigate how it scales to other more complicated problems.

\bibliography{references}
\bibliographystyle{plainnat}
\end{document}